\documentclass{article}
\usepackage{colm2024_conference}
\usepackage{graphicx}
\usepackage{microtype}
\usepackage{hyperref}
\usepackage{url}
\usepackage{booktabs}
\usepackage{multirow}     % For the \multirow command
\usepackage{booktabs}     % For enhanced table formatting
\definecolor{darkblue}{rgb}{0, 0, 0.5}
\hypersetup{colorlinks=true, citecolor=darkblue, linkcolor=darkblue, urlcolor=darkblue}

\title{Multi-BERT: Leveraging Adapters and Prompt Tuning for Low-Resource Multi-Domain Adaptation}

\author{Parham Abed Azad \& Hamid Beigy %\thanks{ Use footnote for providing further information about author (webpage, alternative address)---\emph{not} for acknowledging funding agencies.  Funding acknowledgements go at the end of the paper.} 
\\
Department of Computer Engineering\\
Sharif University of Technology\\
Tehran, Iran \\
\texttt{\{parhamabedazad,beigy\}@sharif.edu}
}

\colmfinalcopy 
\begin{document}

\maketitle

\begin{abstract}
 The rapid expansion of texts' volume and diversity presents formidable challenges in multi-domain settings. These challenges are also visible in the Persian name entity recognition (NER) settings. Traditional approaches, either employing a unified model for multiple domains or individual models for each domain, frequently pose significant limitations. Single models often struggle to capture the nuances of diverse domains, while utilizing multiple large models can lead to resource constraints, rendering the training of a model for each domain virtually impractical. Therefore, this paper introduces a novel approach composed of one core model with multiple sets of domain-specific parameters. We utilize techniques such as prompt tuning and adapters, combined with the incorporation of additional layers, to add parameters that we can train for the specific domains. This enables the model to perform comparably to individual models for each domain. Experimental results on different formal and informal datasets show that by employing these added parameters, the proposed model significantly surpasses existing practical models in performance. Remarkably, the proposed model requires only one instance for training and storage, yet achieves outstanding results across all domains, even surpassing the state-of-the-art in some. Moreover, we analyze each adaptation strategy, delineating its strengths, weaknesses, and optimal hyper-parameters for the Persian NER settings. Finally, we introduce a document-based domain detection pipeline tailored for scenarios with unknown text domains, enhancing the adaptability and practicality of this paper in real-world applications. 
\end{abstract}

\section{Introduction}

\begin{figure}
  \centering
  \includegraphics[width=0.9\textwidth]{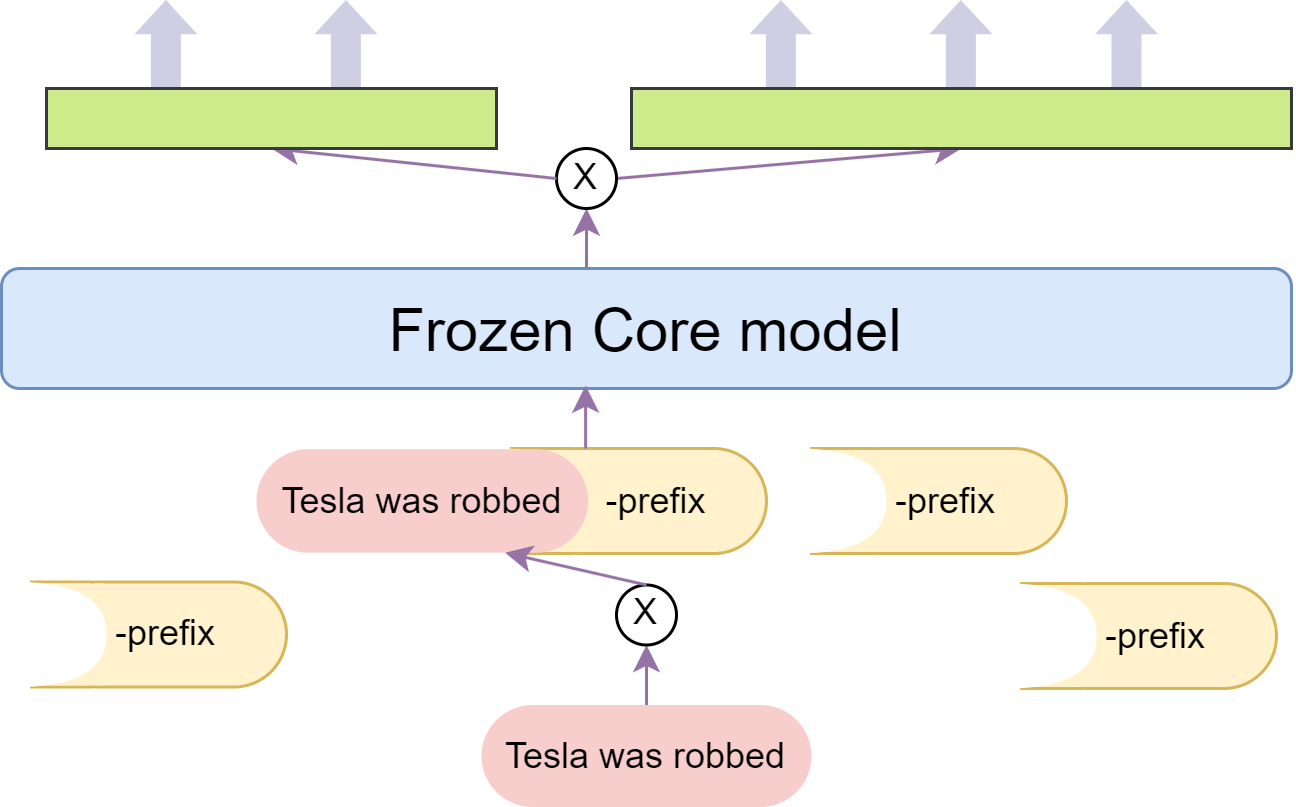}
  \caption{Each set of parameters belongs to exactly one domain but layers are often shared by a couple of domains.}
  \label{fig:model-arch}
\end{figure}
% add huge text = hard we need NER!
Named entity recognition (NER) is an essential part of natural language processing (NLP) that helps in tasks from understanding language and extracting information to question answering and sentiment analysis. Recently, NER has become even more important, thanks to the explosion of interest in NLP, which gave birth to many new challenges. One of the most pressing challenges is the adaptation of NER models to the ever-expanding domains. This task becomes particularly difficult as model sizes and inference times continue to increase.  The dynamic nature of natural languages coupled with the diverse array of topics and contexts spanning different domains, presents a formidable challenge in NLP. As the field advances, it becomes increasingly evident that models trained within a specific textual domain often falter when confronted with data from disparate domains. This discrepancy in performance is exacerbated by the fact that many sentences necessitate varying entity labels depending on the contextual nuances. For instance, consider the sentence "Tesla was robbed": in a scientific or historical context, "Tesla" would likely be tagged as a person, whereas in discussions related to business or economics, or within the context of a casual tweet, "Tesla" would be categorized as an organization. This contextual ambiguity poses one of the primary challenges in accurately identifying entities, particularly within specialized fields such as medicine.~\citep{kundeti2016clinical}.

Traditionally, the preferred approach was to train a single model for all domains. The model is customized in a way that fits all areas. However, there is a paradox with this approach: while we praise these models for performing well without knowing the text domain, these models tend to perform worse than models trained on a single domain simply because of the lack of that domain knowledge. This is because different domains represent completely different situations and may require different responses to the same input. Therefore, no matter how good a trained model is, it will never be able to label the inputs perfectly, unless the model knows the context of the text. This dichotomy has stimulated the exploration of different strategies to overcome the barriers posed by adapting to multiple domains in NER. An additional challenge associated with this approach arises from the limitation of a single model to produce outputs in only one format. Certain domains may require different sets of labels, leading to the necessity for varied output formats tailored to each domain's needs. The adoption of multiple models to accommodate various domains introduces significant drawbacks, including resource-intensive requirements such as extensive RAM and storage constraints, as well as the time-consuming process of training each instance. Moreover, the interconnected nature of different NLP tasks implies that training for one task should inherently contribute to the understanding and performance improvement of another.

To deal with these challenges many have turned to prompt engineering. Models that get the required context from the context of the text. However, while this approach works for some issues it tends to require a huge sum of data to perform competently which we do not have access for many tasks and domains. This problem is extremely exacerbated when we look at the languages that are suffering from a lack of well-labeled and clean data such as the Persian language. We delve deeper into the details of these models in the \ref{sec:related-work} section.

Therefore, a novel approach is proposed. This model offers an innovative solution to address these challenges by leveraging prompt tuning and adapters. Initially, we incorporate specific parameters for each domain and create individual output layers to produce distinct outputs for each domain. Subsequently, the added parameters and the output layers are trained carefully. Each set of these parameters and layers are only used for certain domains. This allows the model to perform perfectly on all domains. Given access to a robust pre-trained model, the core model is frozen; however, it can be also be trained during the training process if a pre-trained model is unavailable. Remarkably, even when facing limited data availability, we observe a significant performance boost compared to other models. Finally, we introduce an innovative approach for situations where the domain is unknown. Our proposal, termed the document-based approach, examines a set of elements to ascertain their domains and then tags each input accordingly. 

The rest of this paper is organized as follows: section \ref{sec:related-work} gives a brief overview of the related projects that try to solve these issues. Thereafter, section \ref{sec:proposed-method} explains the proposed architecture of the multi-Bert model and the model will be thoroughly evaluated in the section \ref{sec:results}. Moreover, section \ref{sec:pipeline} will propose a novel pipeline that deals with the issues that rises from not knowing the exact domain of the texts, while the section \ref{sec:conclusion} concludes the paper.

\section{Related work}\label{sec:related-work}% classify the related works
The landscape of named entity recognition has witnessed a surge in research efforts, particularly with the emergence of prompt tuning and the increased diversity of textual data. Notably, seminal contributions like PUnifiedNER advocate for the adoption of a robust, centralized model capable of cross-domain applicability, accommodating diverse text domains and label sets~\citep{lu2023punifiedner}. By incorporating comprehensive information into prompts and leveraging extensive training data, these model demonstrates a remarkable capability to discern and label various data types with diverse labels. However, the widespread implementation of such approaches faces a significant challenge—the scarcity of relevant labeled data in many instances. This limitation has spurred a shift in focus toward the development of template-free models utilizing few-shot learning. Pioneering studies introduced models that, with a minimal set of labeled examples (typically 16, 32, or 64), can adeptly label inputted text~\citep{wang2022instructionner, lu2023punifiedner,  he2023template, ma-etal-2022-template}. A similar approach will be incorporated into our pipeline. However, we will not need any labeled data as our pipeline works with unlabeled data.

At the forefront of Persian NER, current state-of-the-art models include BeheshtiNER~\citep{taher2020beheshti} and ParsBERT~\citep{farahani2021parsbert}. a BERT model fine-tuned for NER tasks. While the technique of prompt tuning has found extensive application in English NER, the challenges posed by limited resources for the Persian language made the development of single template-free large prompt-tuning models arduous. This challenge extends beyond Persian to various languages and domains, particularly low-resource languages and specialized fields like bioinformatics~\citep{basaldella2017entity}.
In response to these challenges, recent advancements in Persian NER have predominantly focused on fine-tuning models to cater to diverse domains. Outstanding examples of this strategy include ParsTwiNER~\citep{aghajani2021parstwiner}, a BERT model fine-tuned for both informal and formal texts. However, as seen in this paper this model took way more to converge compared to the original Parsbert while still giving much worse results for the formal text. Another example is Hengam, a BERT model tailored for token classification in the tagging of formal and informal texts. As seen, these models exhibit notable drawbacks, such as prolonged training times and diminished performance in previous domains when adapting to new ones. The need for more efficient and adaptable models persists as an ongoing concern within the landscape of Persian NER research.

% write multibert as "multi-Bert"
\section{The proposed method}\label{sec:proposed-method}
To mitigate the challenges  posed by time and size constraints inherent in employing multiple models, a model called multi-Bert is presented. In multi-Bert, a single pre-trained model is completely frozen, while multiple sets of additional parameters and layers are integrated into the model. This configuration allows for the generation of diverse results from a single model. Moreover, this design facilitates training for specific tasks without modifying the underlying base model, thereby safeguarding the performance of one task from affecting another. However, while the domain-specific parameters are completely separate from each other, we use pre-training for each set of parameters on other sets of data. This approach allows us to leverage any correlating information that can aid the model's performance. As seen in \ref{fig:model-arch}, the model allows the selection of task-specific parameters during inference, tailoring the model's behavior to the requirements of each individual task.

A significant advantage of multi-Bert is its efficient use of adapters compared to traditional fine-tuning methods. Adapters accelerate the training process~\citep{he2021effectiveness}, reducing the time required for each epoch and enabling model convergence in fewer than 10 epochs. This efficiency allows us to employ a two-step training approach: first, pre-training one set of parameters on all available data, and then fine-tuning a copy of these parameters for each specific task of interest. By leveraging task-specific data, we can effectively utilize cross-task information during fine-tuning.

To incorporate these parameters, two distinct approaches are adopted: prompt tuning and adapters. After exploring various methods within each approach, the most effective techniques were selected. For prompt tuning, utilize Prefix-tuning has been utilized~\citep{li2021prefix}. Prefix tuning is a technique where a small set of learnable parameters, known as a prefix, is embedded directly into the input of all layers of a pre-trained language model. This allows the model to rapidly adapt to task-specific information without the need to fine-tune all the model parameters. The prefix acts as a continuous task-specific vector that can influence the behavior of the model across all layers, providing a lightweight and efficient way to customize large language models for specific tasks. It has been shown to achieve comparable performance to full model fine-tuning while requiring the tuning of only a tiny fraction of the parameters By leveraging the structured nature of prompts, this approach facilitates prompt-driven learning, a crucial aspect in multi-domain scenarios.

On the other hand, for adapters, we employ the well-known Low-rank adaptation method, also known as LoRA~\citep{hu2021lora}. This method yields comparable results by incorporating learnable parameters into the model layers. LoRA focuses on preserving adaptability without compromising the integrity of the base model. By adding parameters to each layer without introducing new ones, LoRAs have emerged as highly reliable adapters. Their efficiency lies in seamlessly integrating new parameters into existing layers, yielding impressive results within a short time frame, and facilitating straightforward merging of the new parameters with the existing layers.

Additionally, a classification layer is introduced based on the required number of classes. In cases where tasks share the same output structure, both the size of output and the specific labels, this layer can be shared among them. Conversely, for tasks with differing output structures, we accommodate multiple final layers tailored to each task's unique requirements. This streamlined approach not only addresses the challenges associated with multiple models but also provides flexibility in adapting to diverse task requirements. The effectiveness of multi-Bert is validated through comprehensive evaluations utilizing various parameter addition methods and task-specific classification layers.\\
We use a fine-tuned core model, ParsBert which is arguably the best pre-trained Bert model. There are a lot of different models pre-trained for the persian language and each one can be used. However, based on our calculations ParsBert performs the best for the NER tasks. Therefore, ParsBert was used in this study and due to high performance of this model, it was frozen throughout the training steps of the multi-Bert model.

There are a few steps to train the model, firstly, for each domain, exactly one adapter is introduced, and for each set of adapters, that have the same output template, a classifier header is included. Initially, one adapter is trained on all available data associated with that classifier excluding the domain of interest for a couple of epochs to ensure the adapters have all the correlating knowledge from other domains and the classifier is properly tuned. Subsequently, this adapter is replicated across all other adapters of the same classifier and fine-tuned on each domain separately, until convergence and before overfitting.

The classifier is only trained when the adapter is being pre-trained since this step includes all of the data associated with that particular output scheme. Furthermore, this layer is frozen during the fine-tuning of the adapters this really helps make the overall training to be shorter and helps preserve the knowledge of the adapters from inference. After the training process is complete, there is only one core model with multiple headers and many domain specific parameters. Therefore, we have multiple models each tailored for a single domain that are able to perform separately while they share much of their architecture with each other. The subsequent section outlines the implementation and workflow of the proposed model, shedding light on its efficiency and adaptability in handling multi-domain NER tasks. 
\begin{figure*}
  \centering
  \includegraphics[width=1\textwidth]{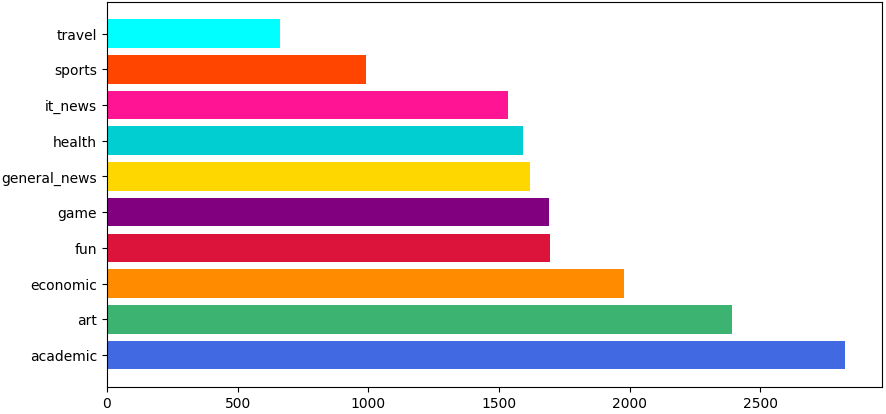}
  \label{fig:dataset}
  \caption{The size of the data in each domain of text greatly differs from one another, which results in massive challenges.}
\end{figure*}
\section{Evaluation and Results}\label{sec:results}  %: A Comprehensive Analysis of Model Performance}
The proposed model is evaluated by using three different distinct datasets, each chosen to represent diverse unique challenges in the realm of NER and to test the performance of our model in dealing when faced with different challenges. We also introduce important baselines to show the effectiveness of our model. Finally, we compare the results of all the models and discuss the hyperparameters and advantages of prefix-tuning and LoRAs.
\subsection{Datasets}
% should I refrence to the dataset even tho it has no papers?
To evaluate the aforementioned strategy the models are tested on three distinct datasets. For the first two datasets, we focus on the classic formal versus informal NER tasks, we utilize the Arman dataset for formal entities, and ParstwiNER~\citep{aghajani2021parstwiner}, for the informal ones. This dataset serves as a benchmark for the adaptability of our model across standard formal and informal contexts characterized by minimal noise. This is very important since in many datasets all the text does fall into these domains. For instance, if we have a close look at the data on Twitter's more established accounts we see that many tweets are written perfectly and cleanly whether in a formal language or an informal one. However, these two datasets are extremely standard, they are both based on the CONLL format, have 21 entity types, and lack considerable errors or use of niche grammar which makes them very similar to the majority of the text the core model is trained on. Hence, the results on these two datasets differ from when datasets with noisy data taken from sites like Twitter are used.

That's where the third dataset, ParsNER~\citep{ParsNER}, comes in, ParsNER, This dataset consists of a huge amount of noise, whether it is words that are tagged inconsistently or general script errors. More importantly, the labels of this model are different than the previous datasets with only nine tags and a "MISC" tag that is supposed to represent any other tag, and probably any other dataset, since it's not based on a standard. The data is taken from posts on Twitter pages reflecting different topics. Thus, the data is clustered and grouped in different domains. These domains are extremely different from each other and as we mentioned at the start of the paper, the tagging will be greatly influenced by the topic at hand as a word like "Iran" is probably a "loc" when we are talking about travel and an "org" when we are talking about economics. This feature turns the differences in domain huge as we will see in later in results that models that are specialized in the domain greatly outperform general models. Moreover, the number of entries in each domain differ from one another, in fact, you can see the number of entries in each domain in the \ref{fig:dataset}. Normally, we would not be able to share one model for this dataset with the previous ones due to these huge differences. Still, with multi-Bert, we will use our model to achieve state-of-the-art results across all of our domains and datasets to show that this model can truly be a solution fit for all problems. Therefore, we have two sets of data with different outputs each one has other domains that we need to focus on and we will show that one instance of our model can give state-of-the-art results across all of these domains and datasets. For more information on the replication of the results you may also check the appendix.
\begin{figure*}
  \includegraphics[width=0.35\textwidth]{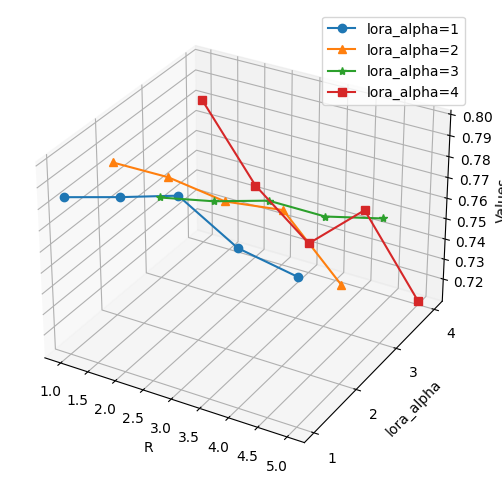}
  \includegraphics[width=0.65\textwidth]{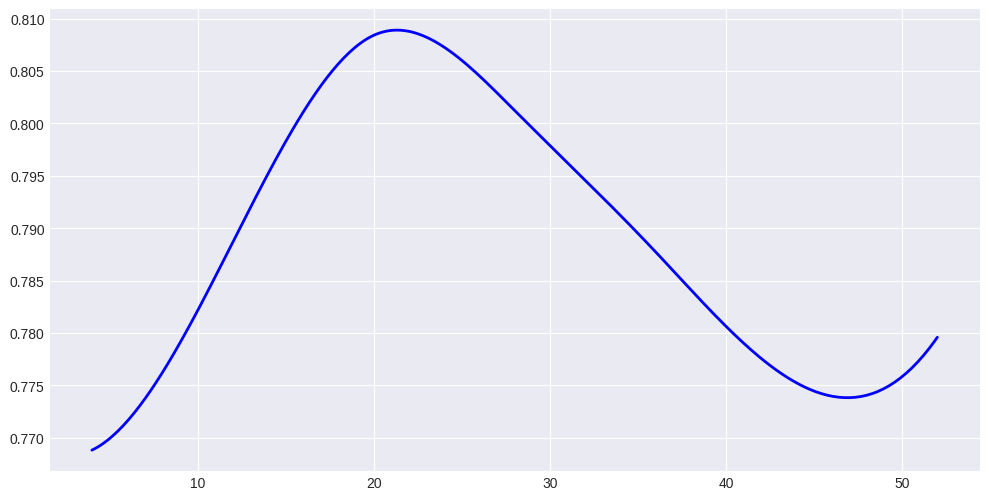}
  \label{fig:hyper}
  \caption{Each line on the left image shows the scores for an specific alpha with different R values. The model on the right however, shows the scores received by the prefix model with different number of tokens added to the inputs of each layer.}
\end{figure*}
\subsection{Baselines}
For the baseline, we introduce two models that we expect our model to perform between them. Firstly, our lower bound baseline is a general model that is trained on all of the domains however since it is impossible for a model to give outputs in two different formats like our multi-Bert we train two general models one for the first two datasets and another for all the domains in the ParsNER dataset. These general models are trained by fine-tuning our pre-trained core model on the concatenation of all domains. However, we also design an upper-bound baseline. We fine-tune the core model on all of our data and fine-tune it on a single domain, we do this twelve times for each single domain. Obviously, this is an extremely time-consuming experiment and the result is twelve huge models that are not really a feasible solution. However, this does give us the best possible solution.
\subsection{Hyper-parameters setting}
One of the main challenges of using prompt tuning or adapters in general is the complex parameters. In this paper, we used an approach that closely resembles grid-search. Firstly we set all parameters to every number that is apart from each other eight(4, 12, 20, e.g) after finding the best performing sets of parameters, we adapt the model to all possible parameters in the range. After many experiments on our different datasets, we came to the final conclusion that the best number of tokens to add to our prompt is eighteen. Adding fewer tokens to each input layer results in worse performance while adding more tokens leads to no positive change in the model performance, if not worse, and only leads to much longer training time. For the LoRA model we also did the same thing but the only difference is that there are two parameters that the result is dependant to the their combinations. Thus,  Moreover, we use a batch size of 16 with a learning rate of 0.0001 as it proven to give the best results.
As we see in the \ref{fig:hyper} the performance of the model rises up to the 18 parameters and starts falling significantly which makes chossing the 18 an easy choice. However, for the LoRA model its a little more complicated due to the fact that the results of the model for each value of R or alpha is depandant on the value of the other one. we came to the conclusion that for our named entity recognition task the best combination is LoRA alpha and r of 1 and 3 respectively
\subsection{Results}
As we see in the final results at \ref{tab:2way} and \ref{tab:10way} the general model under-performs in every domain. The general model trained on the first two datasets performs relatively better than the one trained on the others for multiple reasons. Firstly, the data between the two classes are more even, we see that in the second general model, the domains with much smaller datasets are clearly forgotten for the sake of the bigger domains. Secondly, the more the number of our domains is, the harder it becomes for the model to adapt to all of them. We see that in the results of the domains that have much less data compared to their competitors, the model gets over-fixated on the other domains and greatly under-performs in these domains while doing relatively well in the domains with more data. When we look at the F1 score of the specialized models we see that even though each one of them is trained on all of the data and specialized on a single domain they do not outperform our model by a huge margin, in fact, the multi-Bert with prompt-tuning outperforms fine-tuning a model on multiple instances by a small margin. Another important observation is that prompt-tuning outperforms LoRA and is championed as the best way to create this model. This is due to the problem of limited data. Adapters require more data to train effectively. For this simple reason, we see that LoRA gives us results close to the fine-tuned ones, but is greatly overshadowed by prompt-tuning for the domains with much smaller datasets. 
\begin{table*}[htbp]
\centering
\label{tab:10way}
\begin{tabular}{lcccccccccc} % Adjust the width as needed
\toprule
\multirow{1}{*}{\textbf{Model}}  & \multicolumn{10}{c}{\textbf{Domains}} \\
\cmidrule(lr){2-11}
& acad & art & econ & fun & game & news & med & it & sport & travel \\
\midrule
Gen-bert-9 & 82\% & 66\% & 70\% & 69\% & 78\% & 79\% & 83\% & 86\% & 81\% & 76\% \\
Spec-bert & 86\% & \textbf{87\%} & \textbf{93\%} & 90\% & \textbf{96\%} & 93\% & \textbf{95\%} & 93\% & \textbf{95\%} & 90\% \\
Multi-lr & 70\% & 78\% & 80\% & 86\% & 90\% & 87\% & 86\% & 83\% & 92\% & 87\% \\
Multi-pre & \textbf{90\%} & \textbf{87\%} & 86\% & \textbf{92\%} & 91\% & \textbf{97\%} & 93\% & \textbf{94\%} & 93\% & \textbf{95\%} \\
\bottomrule
\end{tabular}
\caption{The F1 Scores of the general bert is overshadowed in any domain but the difference is much more visible in smaller domains}
\end{table*}
\begin{table}[htbp]
\centering
\label{tab:2way}
\begin{tabular}{lcc}
\toprule
\textbf{Model} & \textbf{Arman} & \textbf{ParstwiNER} \\
\midrule
General-bert-21 & 75.4\% & 95.2\% \\
Spec bert & 81.7\% & \textbf{99.6}\% \\
Multi-bert-lr & 80.8\% & 99.4\% \\
Multi-bert-pre & \textbf{83.1\%} & 99.3\% \\
\bottomrule
\end{tabular}
\caption{The models all manage to reach closer results this is partly due to the similarity of the text to the core model's training data. }
\end{table}
\subsection{Visual results}
\begin{figure*}
  \centering
  \includegraphics[width=0.8\textwidth]{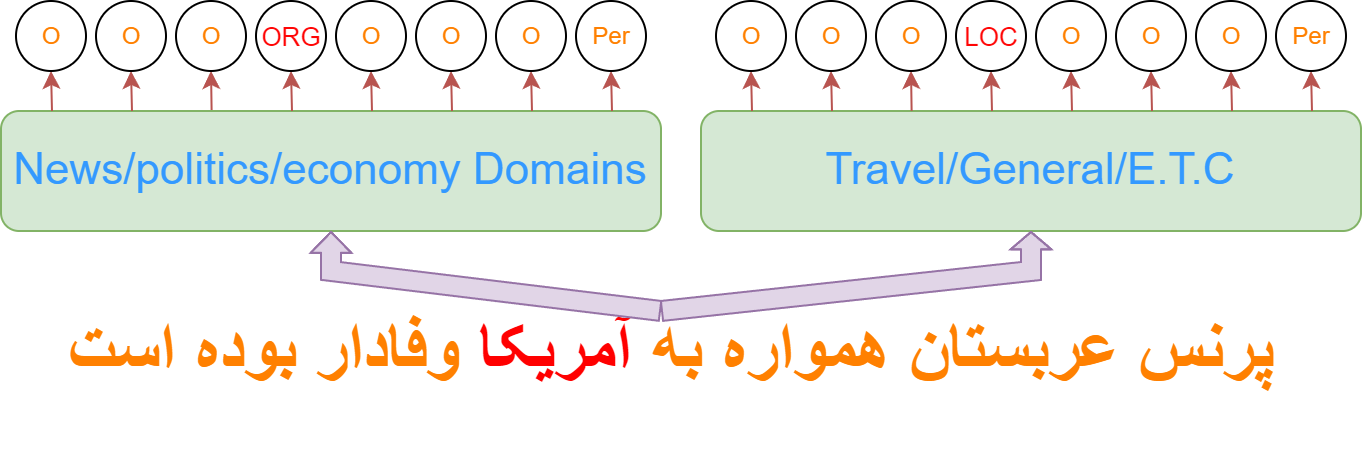}
  \label{fig:example}
  \caption{The models specialized on News, politics and economy get the tag of US right. Text from the political domain of ParsNER. Text translation: The The Saudi prince proved to be loyal to the United States.}
\end{figure*}

So the model is getting better results, but why and where are these improvements? To answer these essential questions in this section the domain models are tested on a particular example from the ParsNER dataset in the general news section. The translation of the sentence goes "The Saudi prince proved to be loyal to the United States". The word United States has a huge ambiguity here, if the loyalty is to the land of the country it needs to be labeled as "LOC", however, if the point is to be loyal to the government of the country the label would be "ORG". To us, humans, labeling this sentence might not be that hard, after all from the tone and the context we might be able to understand that the context of the sentence is politics. But this sentence proves to be exceptionally hard for the model, as seen in the \ref{fig:example} not only do the travel domains get this answer wrong but general models and others such as the ones designed for IT news also get this sentence wrong while the models that know the context of politics, economics or even general news get it right. This also outlines that a small boost goes a long way as seen in some specialized models.
\section{Document-based classifier pipeline} \label{sec:pipeline}
Throughout this paper, we have operated under the assumption that we are aware of the specific domain we are dealing with. However, in this section, we address the challenges posed by the absence of domain knowledge and propose an innovative solution to overcome this obstacle. Fortunately, determining the domain of a given text becomes relatively straightforward when the text is sufficiently long or when multiple samples are available. Nevertheless, feeding multiple samples simultaneously to a model is impractical, as it may lead to unwanted interference among distinct entries. To tackle this issue, we introduce an independent context model by fine-tuning a new set of parameters to our core model. This enables us to adapt the model to classify each example using the existing data.

\begin{figure}
  \centering
  \includegraphics[width=0.8\textwidth]{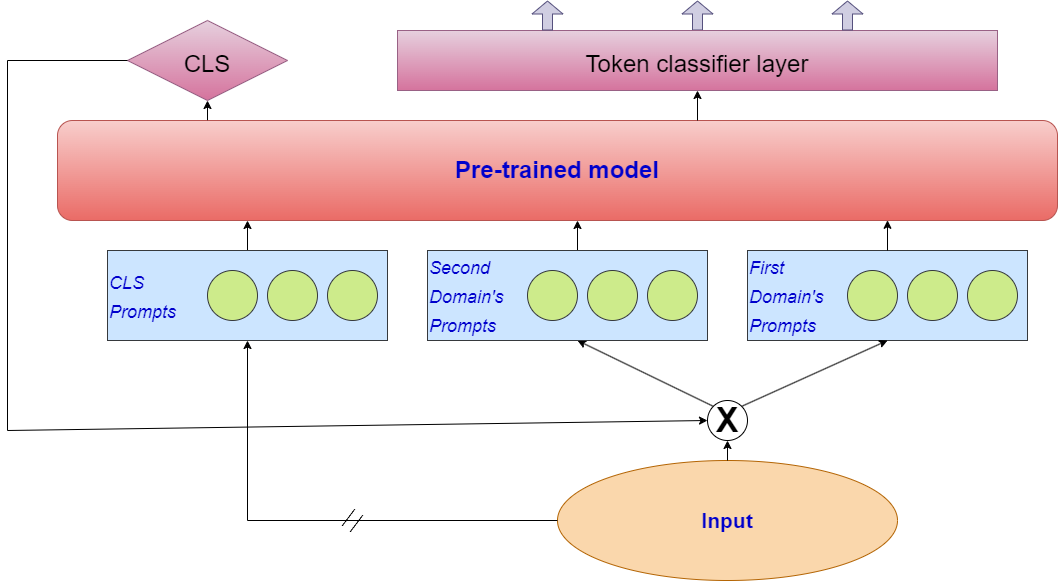}
  \label{fig:pipeline}
  \caption{One forward pass determines the domain of a set which then can be used for each single input.}
\end{figure}

To achieve this objective, we aggregate every set of elements (e.g., 8 elements in this experiment) and assign them a label representing the domain of the data. Subsequently, we amalgamate and shuffle the data from all domains and train the model with the additional parameters tailored for the classification task. Upon completion of the training process, we construct our pipeline as seen in the figure \ref{fig:pipeline}. When employing the model for inference—whether it involves tagging a series of comments on a website, tweets within a Twitter thread, or processing a lengthy book—we provide 512 tokens from the text to the model while utilizing the classification adapter or prompts. Subsequently, based on the identified domain, we apply the respective parameters from the core model to obtain the final results.

It is important to note that since we utilize the core model already employed in our token classification tasks and entirely freeze the core model during the classifier training, this pipeline does not adversely impact the main token classification models. To train, we concatenate each 8 input rows as one input. However, we only concatenate up to a length of 512. Therefore, if the sum size of 5 elements exceeds 512 we only concatenate 5 elements and cut the first 512 tokens of it. Consequently, each group of inputs turns into one input with the label of the dataset they are picked from. Then, we mix and shuffle all of the data and train and evaluate the model on all the data.
To keep the integrity of evaluations, this method is not used in any of the previous experiments. However, here we will run tests to see how this pipeline performs on the mentioned datasets. Firstly, For the formal and the informal datasets, we get an astonishing accuracy of 100\%. This is very reasonable when we pay attention to the distinctions between these two datasets. However, when we look at the ParsNER dataset, with all of its' domains, and we want to do the same there we see less desirable results with an accuracy of 97\% which is not that bad of an accuracy since we have 8 different classes. Moreover, even if this model fails to predict the right domain of the text, it does not guarantee a wrong output as the chosen version may generate correct results. In fact, the decided domain is probably extremely close for this mix-up to happen. Hence we can use this pipeline to use the collection of texts to decide on their context and then process them normally one by one.

\section{Conclusions}\label{sec:conclusion}
This paper proposed the multi-Bert model. This model is designed to perform exceptionally for all domains with any set of outputs. This is thanks to the deliberate design of this model by adding parameters for each domain and output layers for different sets of outputs coupled with the faster training time. This design allowed the model to perform the task on all domains perfectly. Moreover, this paper evaluated the proposed model on Arman (a formal dataset), ParstwiNER (an informal dataset), and the ParsNER, a collection of ten datasets from different contexts with large amounts of noise. The results proved that this model performs as well as the state-of-the-are for each domain, if not better. In addition, we also proposed a pipeline that can decide the domain of the data when a small set of sentences are available. We observed that if we use the model for sets of 8, it could understand the formality of the inputs completely with a 100\% accuracy and can classify the exact domain of the news with an astonishing accuracy of 97\% for the ParsNER dataset when it had access to only 8 samples at a time. 

There is much to do in the future as this paper is only one step toward dealing with multi-domain problems. First and foremost, there should be an adapter that is designed to work for multi-domain settings. An adapter that can understand different domains while sharing needed parameters and can distinguish domains with different output layers. Secondly, this approach should also get tested for other low-resource NLP tasks, especially in bioNLP tasks where there is a huge set of domains. Last but not least, there is much room to improve the core model in a way so that it's adaptable much easier and faster. However, this is not as easy as it might seem initially. This is because we do not want the model to adapt to new knowledge without fine-tuning it. Therefore, if the model is trained on all of the domains it has completely missed the point of adaptability. A simple way to achieve this is to train the model on the sum of losses of the model. However, there must be much research to find the most elegant way to deal with this problem. No matter where the future works take us, there is much room to improve, and this paper poses an incredible stepping stone for future endeavors.

\bibliography{custom}
\bibliographystyle{colm2024_conference}
\end{document}